\documentclass{article}

\usepackage[preprint]{corl_2021} 

\usepackage{epsfig}
\usepackage{graphicx}
\usepackage{amsmath}
\usepackage{amssymb}
\usepackage{mathptmx}
\usepackage{mathrsfs}
\usepackage{xcolor}
\usepackage{makecell}
\usepackage{mathtools}
\usepackage{bbm}
\usepackage{booktabs}
\usepackage{microtype}
\usepackage{xspace}
\usepackage{multirow}
\usepackage{soul}
\usepackage{tabularx}
\usepackage{enumitem}
\usepackage{comment}
\usepackage{float}

\usepackage{wrapfig}
\usepackage{caption}
\usepackage{subcaption}
\usepackage{amsmath}

\DeclareMathSizes{10}{10}{9}{4}

\newcommand{\mypara}[1]{\vspace{1mm}\noindent \textbf{#1}}
\newcommand{\titlecap}[2]{\textbf{#1} #2}

\title{Single RGB-D Camera Teleoperation for General Robotic Manipulation}

\author{
  Quan Vuong, Yuzhe Qin, Runlin Guo, Xiaolong Wang, Hao Su, Henrik Christensen \\
  University of California San Diego \\
  \texttt{\{qvuong, y1qin, ruguo, xiw012, haosu, hichristensen\}@ucsd.edu} \\
}

\begin{document}

\vspace{-1em}
\maketitle
\vspace{-2em}



\begin{abstract}
    We propose a teleoperation system that uses a single RGB-D camera as the human motion capture device. Our system can perform general manipulation tasks such as cloth folding, hammering and 3mm clearance peg in hole. We propose the use of non-Cartesian oblique coordinate frame, dynamic motion scaling and reposition of operator frames to increase the flexibility of our teleoperation system.  We hypothesize that lowering the barrier of entry to teleoperation will allow for wider deployment of supervised autonomy system, which will in turn generates realistic datasets that unlock the potential of machine learning for robotic manipulation. 
    Demo of our systems are available online \footnote{\url{https://sites.google.com/view/manipulation-teleop-with-rgbd}}.
    \end{abstract}


\vspace{-1mm}
\section{Introduction}
\vspace{-1mm}
\label{sec:intro}

In machine learning, training data should approximate the test distribution. Not surprisingly, a key ingredient in the successes of learning algorithms is diverse and realistic datasets. The collection of such datasets is made possible by existing platforms whose main purpose is not dataset collection. For example, without the internet, it would have been prohibitively costly to collect the $22$ million images in ImageNet at the same level of visual diversity. Similarly in NLP, language models are trained from large corpus crawled from the internet~\cite{GPT3}, ensuring syntax and semantic diversity. In summary, existing data collection platforms have enabled rapid progress in both CV and NLP.

A key property of these platforms is that the collection of diverse datasets is \textit{only} a by-product of deploying socially useful services. The immediate values to society offered by these platforms offsets the cost of large-scale data collection. From this perspective, a key bottleneck preventing the progress of learning for manipulation is the lack of a data acquisition mechanism \textit{in the wild}, such as in actual homes and offices. In fact, we are faced with a chicken-and-egg problem. From a learning perspective, developing generalizable manipulation robots relies on large-scale and realistic datasets. However, collecting such datasets requires deploying robots to perform tasks in the real world. How can we break this cyclic dependency? How do we develop semi-autonomous robots that can perform useful tasks in realistic environment conditions, thereby generating realistic datasets as a by-product?

We hypothesize that supervised autonomy~\cite{SupAutoTeleOp,SupAutoHRI} through remote teleoperation will play a key role. In supervised autonomy, the robots do not need to generalize well to all operating conditions. A human-in-the-loop can take over control of the robot when the autonomous system is expected to fail. Current teleoperation interfaces usually require special hardware that is not readily available, such as Virtual Reality headset controller or game controller. We are interested in reducing the barrier of entry to teleoperation. Specifically, we are interested in the following question:

\textit{
Given only one RGB-D camera as the motion capture device, to what extent can a trained human operator command a robotic arm to perform complex manipulation tasks? 
}

\begin{figure*}[!t]
\setlength{\tabcolsep}{1pt}
\begin{tabular}{cccc}
  \includegraphics[width=0.22\linewidth]{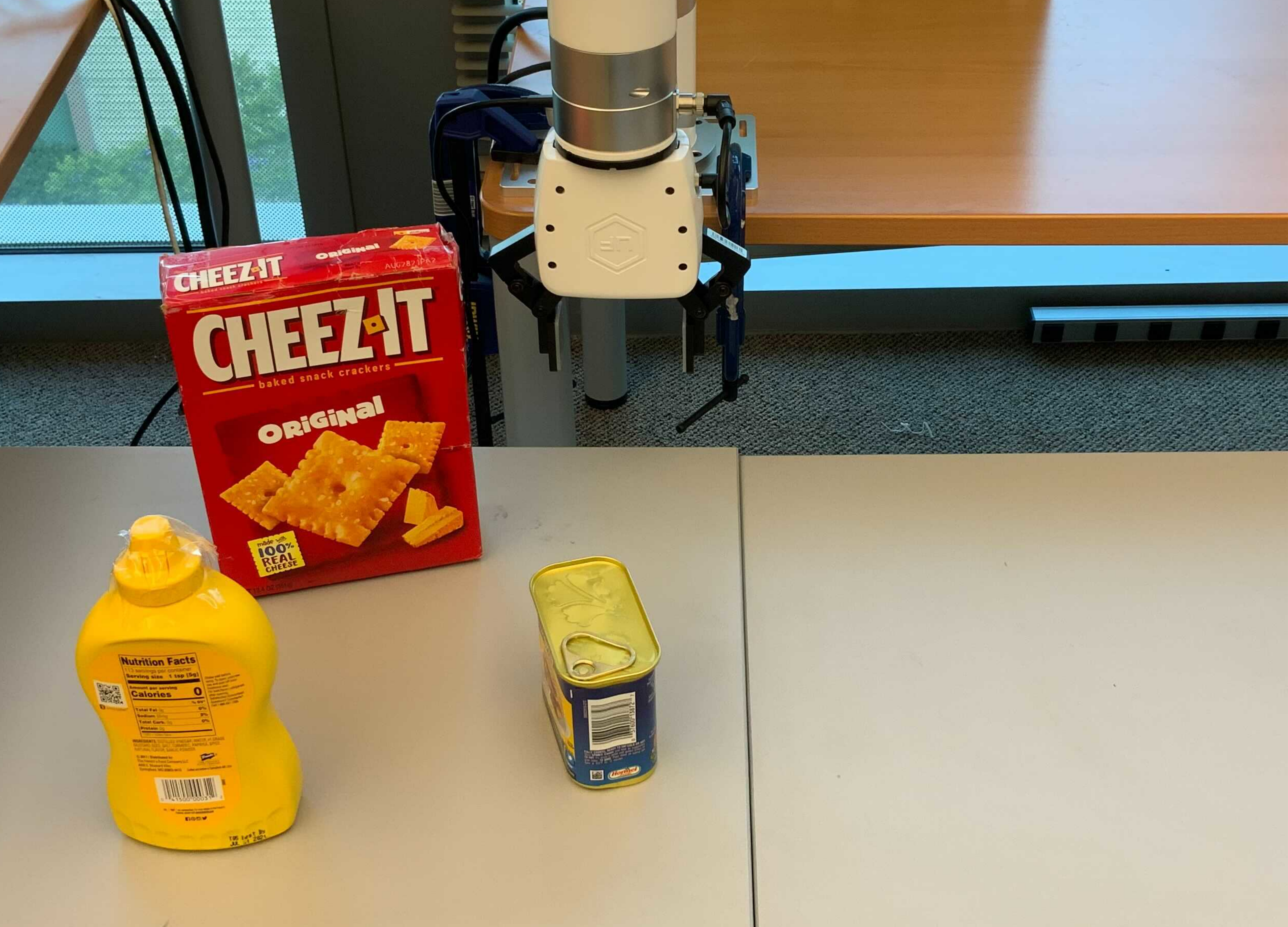} &   \includegraphics[width=0.22\linewidth]{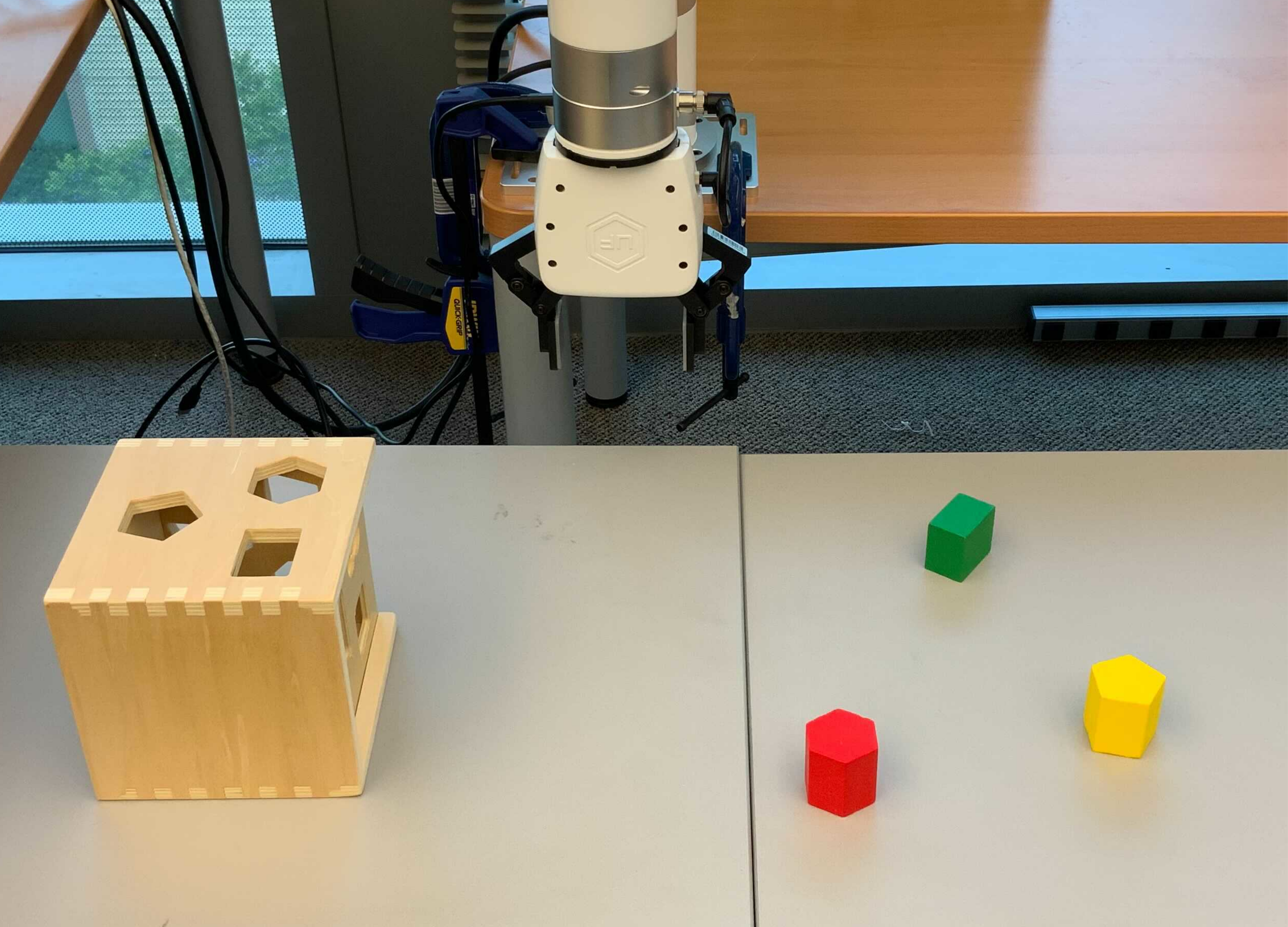} & \includegraphics[width=0.22\linewidth]{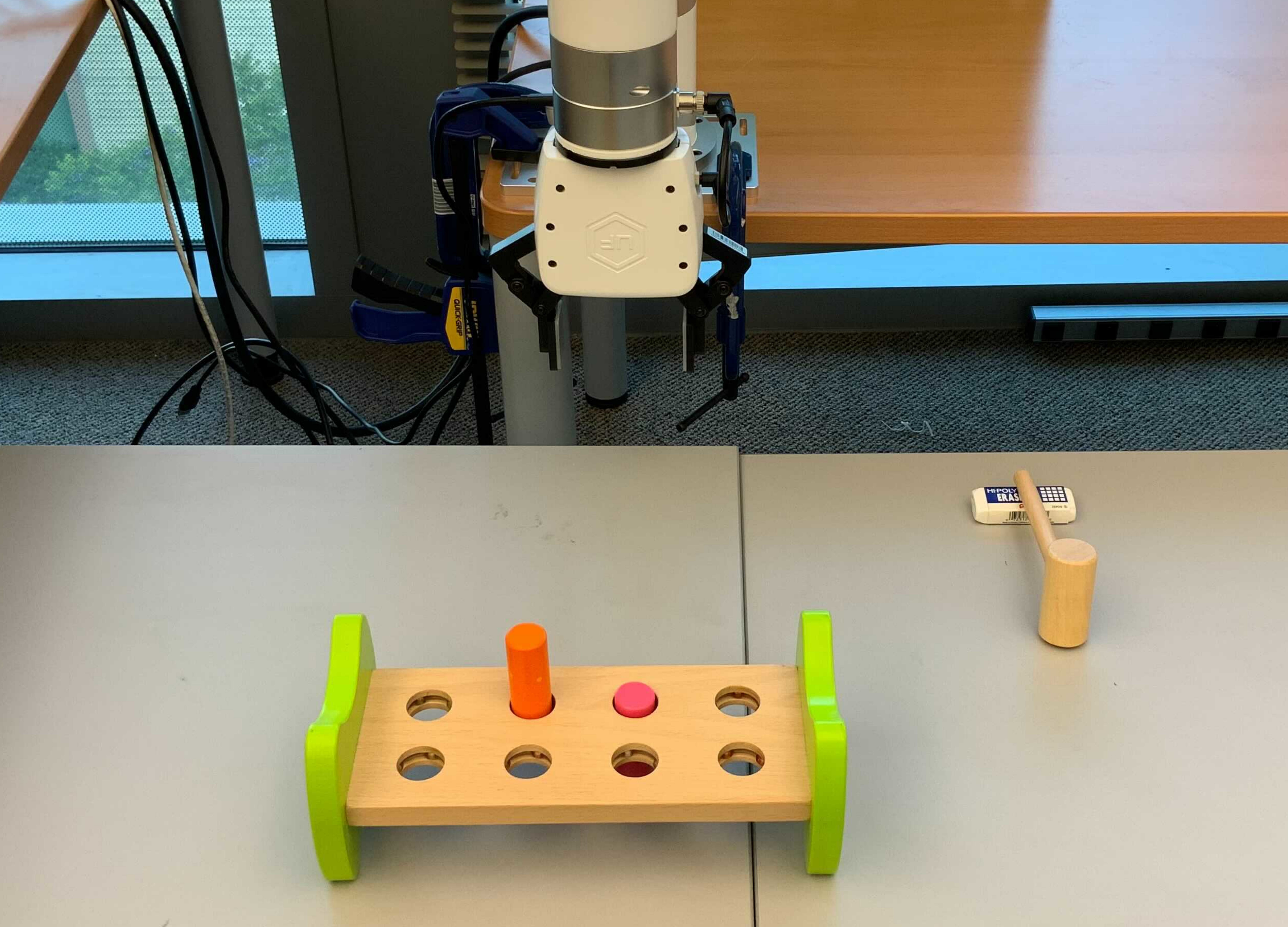} &   \includegraphics[width=0.22\linewidth]{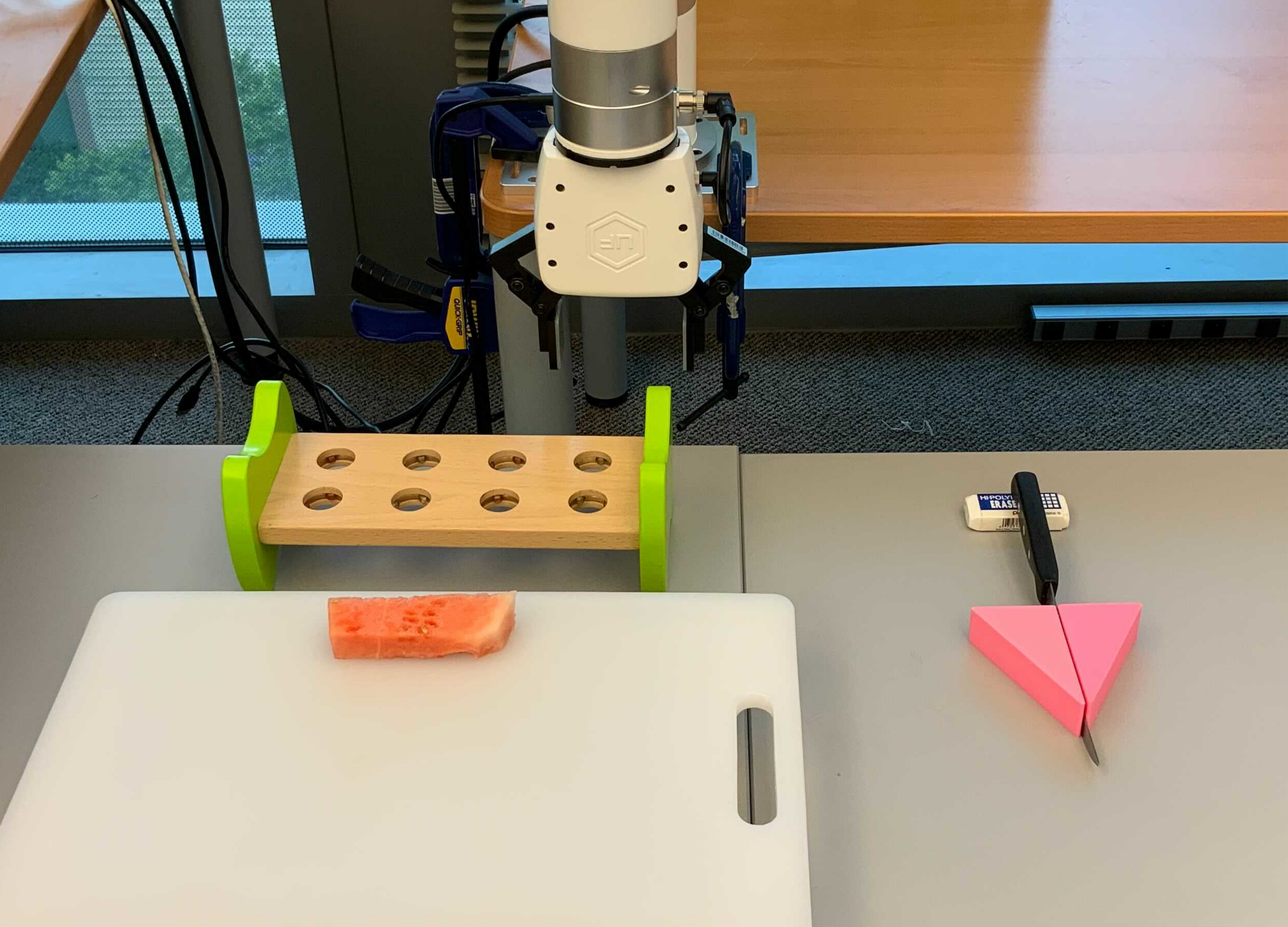} \\
(a) Pick and Place & (b) Peg in Hole & (c) Hammering & (d) Cutting Fruit \\[6pt]
  \includegraphics[width=0.22\linewidth]{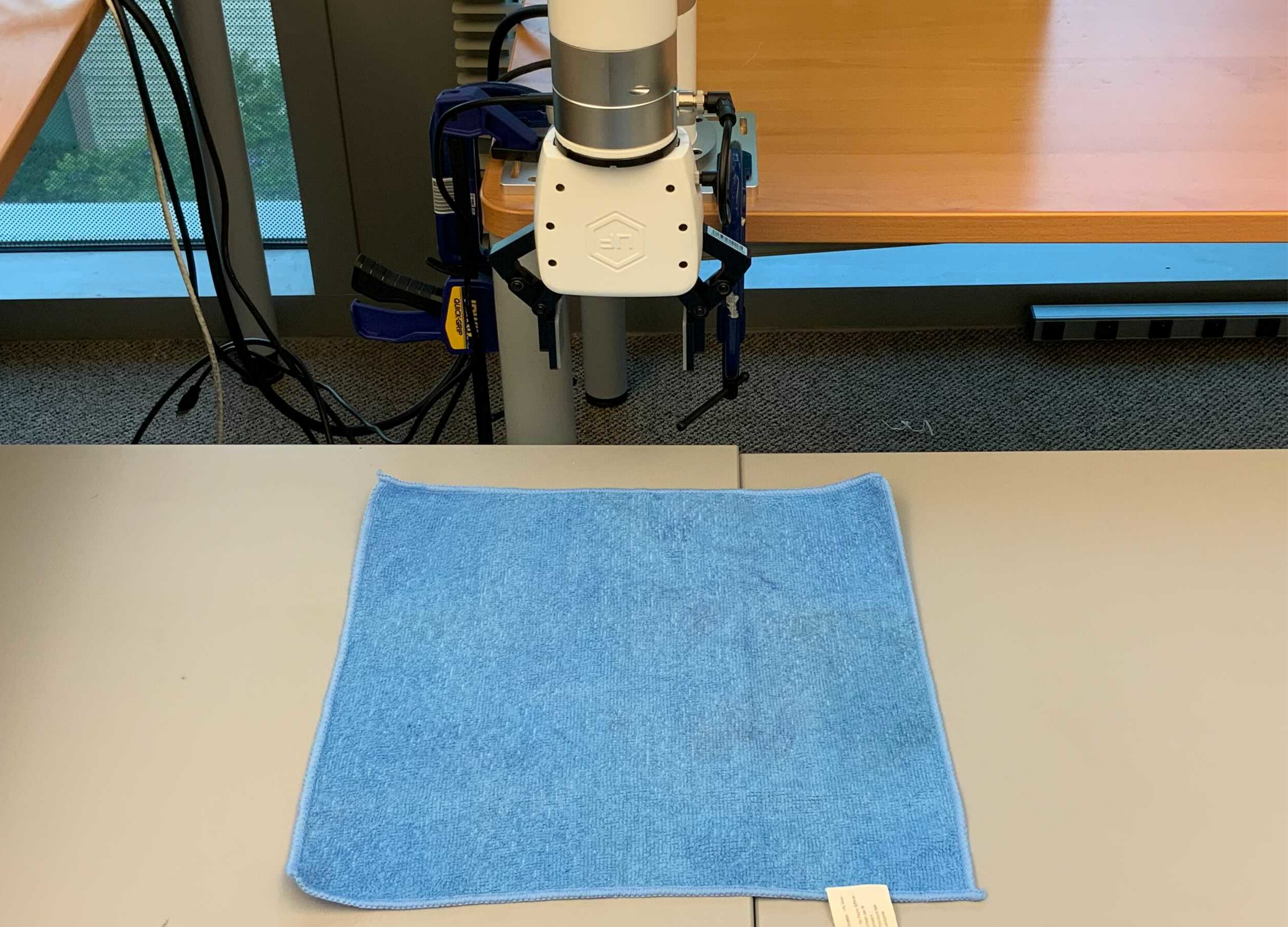} &   \includegraphics[width=0.22\linewidth]{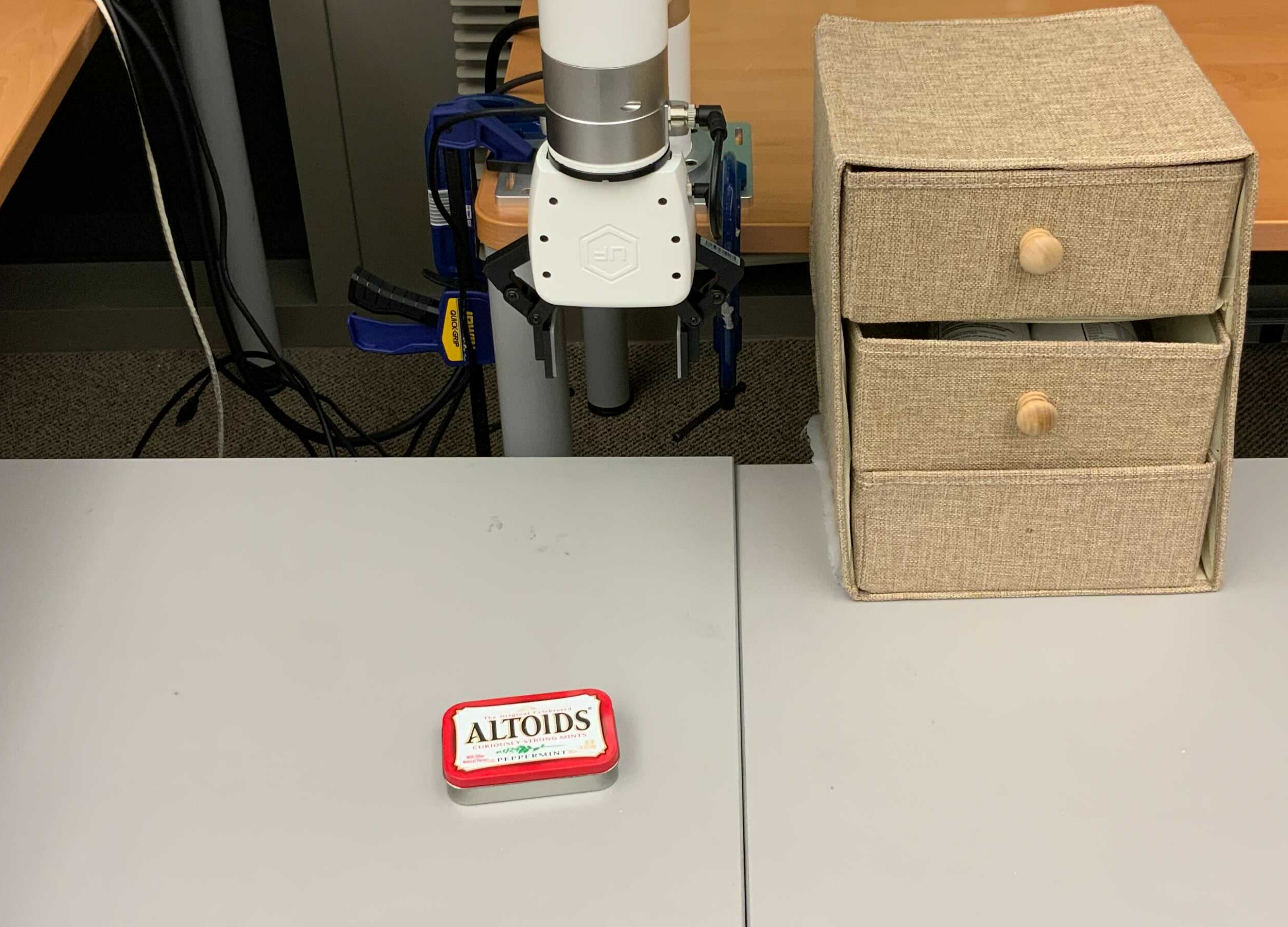} & \includegraphics[width=0.22\linewidth]{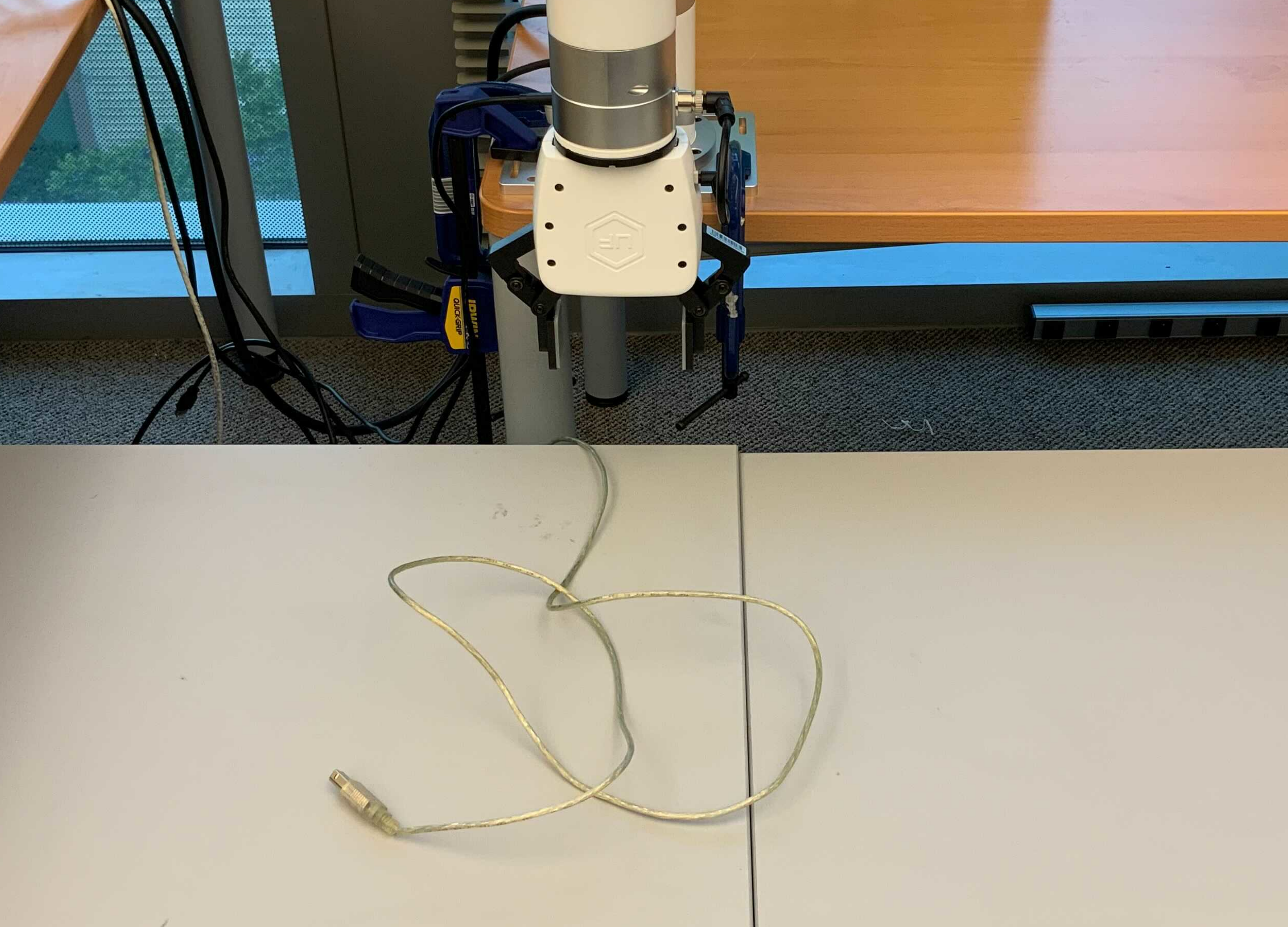} &   \includegraphics[width=0.22\linewidth]{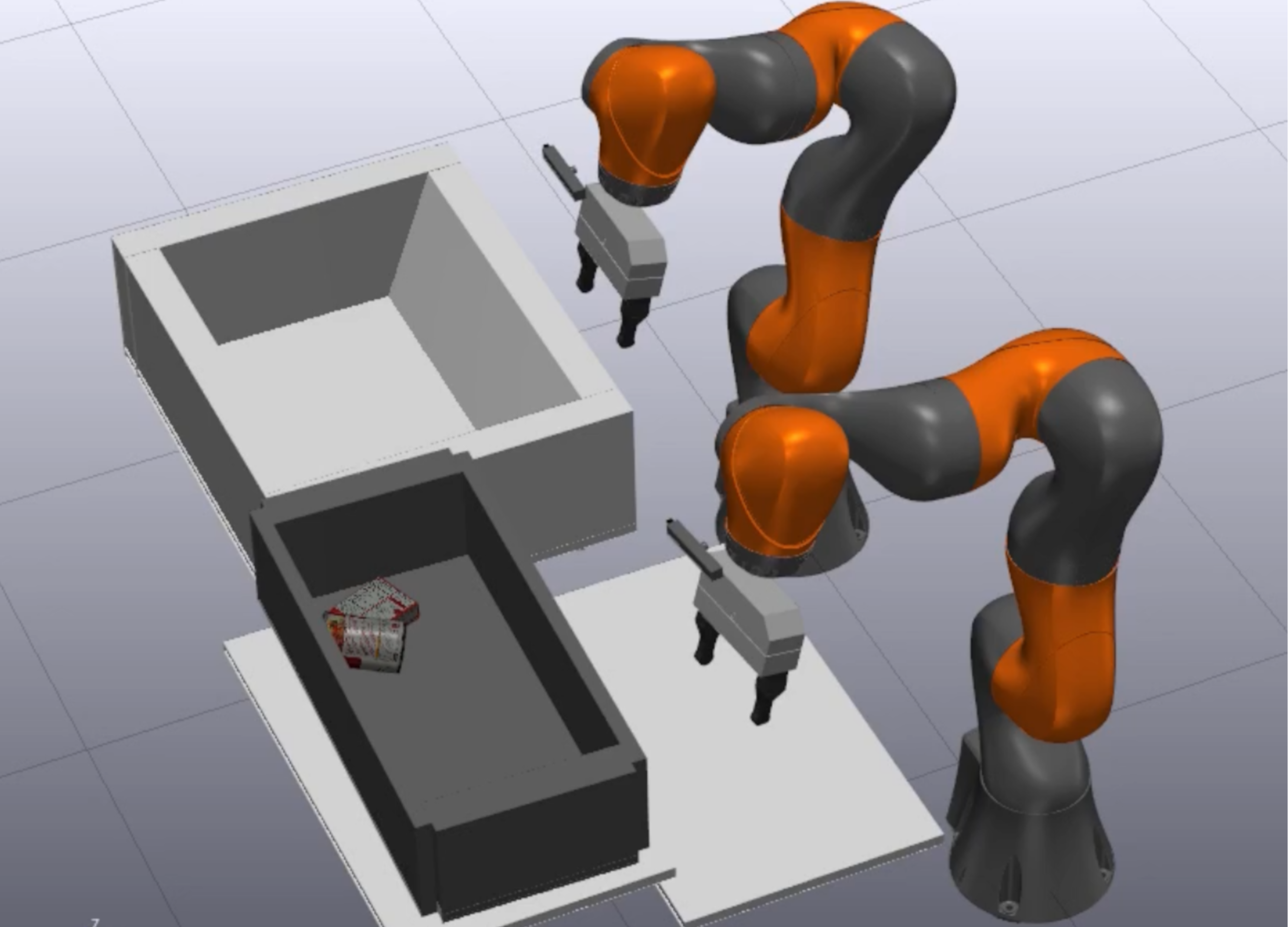} \\
(e) Folding Cloth & (f) Object Storage in Drawer & (g) Cord Untangling & (h) Moving Large Container\\[6pt]
\end{tabular}
\caption{We propose a teleoperation system which uses a single RGB-D camera to capture human motion. The human operator does not have access to any form of tactile feedback. Tasks (a)-(g) are evaluated on a real robot arm. Dual arm task (h) is performed in simulation. \vspace{-4mm}}
\label{fig:teaser}
\end{figure*}

Figure~\ref{fig:teaser} illustrates the tasks we perform with a single camera. 
Our contributions are summarized as:
\begin{itemize}[leftmargin=10pt]
    \item A teleoperation system for manipulation which uses a single RGB-D camera for motion capture.
    \item Novel application of non-Cartesian oblique coordinate frame to define the operator frame, dynamic motion scaling, and repositioning of operator frame for more intuitive control.
    \item Demonstration that our system can successfully accomplish complex manipulation tasks.
\end{itemize}
To the best of our knowledge, in teleoperation, we are the first to demonstrate the successful completion of several complex manipulation tasks while using only a single RGB-D camera as sensing device. We also show preliminary results on removing depth sensing while still allowing the operator to command the robot by performing monocular depth estimation. 

We emphasize the mutually beneficial relationship between ours and orthogonal research directions, such as developing physically accurate simulator \cite{xiang2020sapien} or RL algorithms that learn from offline data \cite{DBLP:journals/corr/abs-2104-07749, DBLP:journals/corr/abs-1909-11373}. For example, if we are successful in reducing the barrier to entry of teleoperation and in turn enable large-scale data collection \textit{in the wild} for robotic manipulation, the resulting datasets can possibly be used to improve the fidelity of physical simulators~\cite{DomainRandSim2Real,SimulationRandSim2Real,AutoTuneSim2Real}.

The organization of the paper is as follows. We discuss related work in \autoref{sec:related}. \autoref{sec:task} and \autoref{sec:method} illustrate the hardware and software design of our teleoperation system. \autoref{sec:problem} describes our experimental results and we end the paper in \autoref{sec:conclusion}.

\vspace{-1mm}
\section{Related Works}
\vspace{-1mm}
\label{sec:related}

\mypara{Manipulation Dataset Collection}
The need for large-scale datasets is well recognized within the robot learning community. There have been various efforts to collect real robot data in lab settings \cite{MIME,yu2016more,chebotar2016bigs, ROBONET} or to develop low-cost control interface for simple tasks such as grasping \cite{GraspInTheWild}. We pursue a different philosophy and aim to lower the cost of teleoperation to allow for real world deployment. There have also been efforts to scale up data collection by taking advantage of human videos \cite{VisualImitation} or simulations \cite{dapg, MultiArmTeleop, DBLP:journals/corr/abs-2012-06733}, which face their own domain shift challenges when transferred to real robots. \cite{ROBOTurk} proposes to use phone's IMU as the motion capture device to control a robot arm. We demonstrate that our system can use a single RGB-D camera to control two robot arms and perform a task that necessitates two arms for successful execution. We also test our systems on complex tasks that require distinctively different skills to accomplish.

\mypara{Marker-based motion capture and teleoperation.}
Motion capture system tracks human motion and is the key ingredient for robot teleoperation. 
Pioneering motion capture works use special-purpose markers or gloves to track the 2D or 3D positions of critical points~\cite{lamberti2011real, dorner1994chasing, theobalt2004pitching, fang2017robotic, cerulo2017teleoperation}. The tracking results can be further combined with Virtual Reality devices for better feedback to human operators~\cite{wang2009real}. The marker-based solution often suffers from self-occlusion, thus redundant markers are often placed on the gloves to improve tracking consistency. Additional hardware, such as inertia measurement unit~\cite{li2020mobile, miller2004motion}, electromyography \cite{luo2019teleoperation}, and haptic devices \cite{kumar2015mujoco}, can be used to further improve tracking performance. Tactile feedback \cite{broeren2004virtual, jain2019learning, johansson2009coding} to the human operator provided by specially-design gloves also improves the controllability of the teleoperated robots. While these systems offer good teleoperation performance, they require the human operator to have access to hardware that are not widely available, and are thus difficult to deploy widely.

\mypara{Vision-based motion capture and teleoperation.}
Due to their less stringent hardware requirement, vision-based markerless hand tracking has received significant attention from the teleoperation community. ~\cite{sharp2015accurate, li2019vision, antotsiou2018task} tracks the hand poses via monocular camera or depth sensor. A hand model can be also be used to improve tracking performance \cite{kofman2007robot, du2012markerless}. These works use the hand pose tracking results to control a robot arm with parallel gripper for simple tasks, e.g. pick and place. In contrast, our teleoperation system can perform complex tasks such as cutting and cloth folding. DexPilot \cite{handa2020dexpilot} also demonstrates successful completion of complex tasks using a multi-finger hand. To control the multi-finger hand, DexPilot requires high tracking accuracy of most finger joints. The human operators thus require access to four RGB-D cameras with well-calibrated extrinsic to register the complete points of human hand. The operators are also constrained to move within a limited workspace above a black-clothed table to command the robot. The human operator in our teleoperation system can perform complex tasks while using only 1 RGB-D camera without extrinsic calibration. We however recognize that the multi-fingered setting studied in DexPilot is significantly more challenging than our parallel yaw setting. 

\mypara{Hand pose estimation.}
Both marker-based and vision-based teleoperation system make use of hand pose estimation techniques, which have advanced rapidly due to the integration of deep learning and hand models. Motivated by the success in body pose estimating of blend skinning techniques~\cite{loper2015smpl, kavan2005spherical}, MANO~\cite{romero2017embodied} models shape variation from hand scans by learning pose dependent blend shapes and represents the hand as a combination of shape parameters and pose parameters. ~\cite{boukhayma20193d, liu2021semi, rong2020frankmocap} also adopt linear blend skinning model for hand pose estimation and tracking. Our teleoperation system illustrates that state-of-the-art vision-based hand pose estimation techniques are accurate enough to allow for the completion of tasks that require mm-level precision, such as peg-in-hole. 

\vspace{-1mm}
\section{Hardware System}
\vspace{-1mm}
\label{sec:task}


\begin{figure*}[t!]
    \centering
    \includegraphics[width=0.9\linewidth]{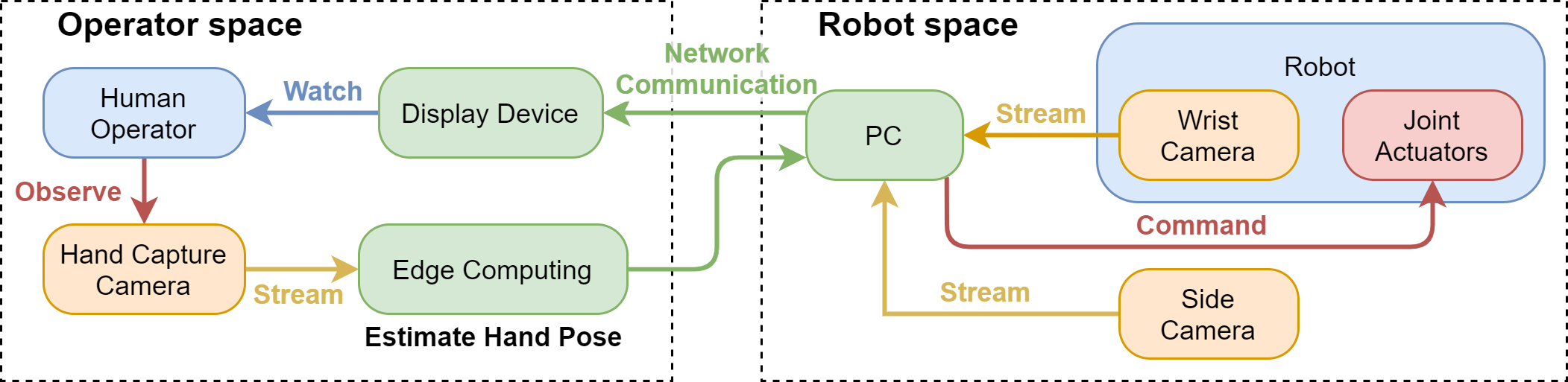}
    \caption{Schematic view of our hardware system.\vspace{-2mm}}
    \vspace{-1em}
    \label{fig:hardware_system}
\end{figure*}

Figure~\ref{fig:hardware_system} illustrates the schematic view of our hardware system. The operator observes the current status of robot workspace, such as the state of the robot and the objects, via a display device in real-time. The operator can move their hands to control the motion of robot arm and grippers. The stream of human motion as captured by the RGB-D camera is processed by an edge computing device to estimate the hand poses. The edge computing device then sends a position command to the PC in the robot space, which is then executed by the robot controller.

There are two cameras in the robot space: a wrist-mounted camera to capture detailed interaction between the grippers and objects, and a tripod-mounted camera to observe the full view of robot workspace. The RGB video from the two cameras are streamed to the display device in the operator space. We use the built-in controller provided by the manufacturer. It already takes count of self collision and collision with the mounting table thus no further motion planning is required.


\vspace{-1mm}
\section{Software System}
\vspace{-1mm}
\label{sec:method}

\begin{figure*}[t!]
    \centering
    \includegraphics[width=\linewidth]{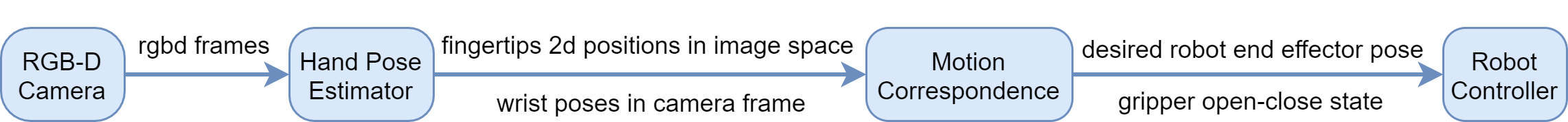}
    \vspace{-4mm}
    \caption{The teleoperation system processes RGB-D frames from the operator space camera and sends end effector pose command to the robot's controller.}\vspace{-2mm}
    \label{fig:software_system}
\end{figure*}

Figure~\ref{fig:software_system} illustrates the inputs and outputs of the software components in our teleoperation system. The three components are: hand pose estimator, motion correspondence, and robot controller. The following sub-sections describe them in more details. We use the distance between right thumb and index tips to determine when the gripper should close \cite{du2012markerless, kofman2007robot}. We will use \textit{monogram} notation in the following subsections to describe frame transformation~\cite{Drake_Notation}.

\subsection{Hand Pose Estimator}

The hand pose estimator takes as input RGB-D frames and outputs $21$ 2d positions of keypoints in image space and the wrist pose in camera space for each hand. Given the RGB frame, we use MediaPipe \cite{zhang2020mediapipe} to estimate the 2d positions of hand keypoints. Given the keypoints, we compute axis-aligned bounding-box to crop the image around hand region. The cropped images are then input into a hand pose regressor proposed by Frankmocap~\cite{rong2020frankmocap} to predict the orientation of the wrist in camera space. We also obtain the 3d positions of the wrist in camera space by deprojecting from the depth map. The hand pose estimator is implemented as a ROS~\cite{ROS} node with publish rate of $10Hz$.   


\subsection{Motion Correspondence}
\label{sec:operator_frame}



Given the estimated hand poses, the motion correspondence module uses a define-at-runtime operator frame to transform the current hand poses into the desired end effector pose. In this sub-section, we motivate the need for the operator frame, introduce its Cartesian and non-Cartesian oblique instantiations and describe how they are used to find the desired end effector pose.

\mypara{Why do we need operator frame?} An issue that frequently appears in teleoperation is viewpoint mismatch. Human operators naturally interpret their motion commands to the robots with respect to an egocentric frame. 
If the frame in the robot space with which the desired end effector pose is represented with respect to (wrt) is not aligned with the egocentric frame, the operators need to understand the relationship between their egocentric frames and the robots' frame and perform mental rotation during teleoperation \cite{kozhevnikov2001dissociation, stransky2010mental, meneghetti2017role}. Such calculation severely impedes the ease of teleoperation. To solve the viewpoint mismatch problem, our system allows the operator to define an egocentric frame that naturally aligns with the frame used for interpreting the end effector control command to the robots. We refer to such an egocentric frame as an operator frame.

\begin{figure*}[t!]
    \centering
    \includegraphics[width=\linewidth]{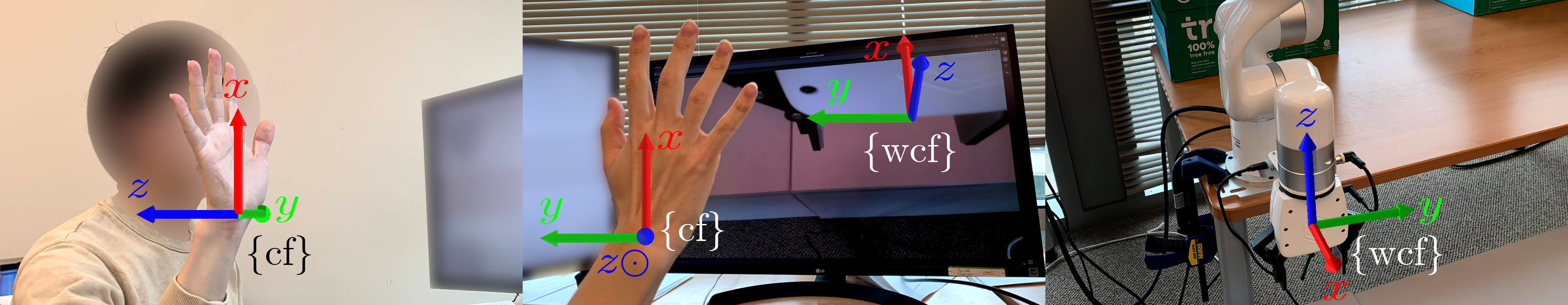}
    \caption{\titlecap{Illustration of the Cartesian operator frame and the  wrist camera frame.}{
    The left and middle figures illustrate the axes of the Cartesian operator $\{ cf \}$ as seen from the operator space camera and by the operator respectively. The middle and right figures illustrates the wrist camera frame $\{ wcf \}$ as seen by the operator and in the robot space. The rough alignment between the operator frame $\{ cf \}$ and camera frame $\{ wcf \}$ from the perspective of the operator, as shown by the middle figure, eliminates the viewpoint mismatch issue.}
    }
    \label{fig:illustration_of_operator_frame}
    \vspace{-0.5em}
\end{figure*}

\mypara{Constructing the Cartesian operator frame.} To solve the viewpoint mismatch issue, in our teleoperation system, the desired robot end effector pose is represented wrt a robot wrist camera frame $\{ wcf \}$. The robot wrist camera frame is illustrated in \autoref{fig:illustration_of_operator_frame}. Our system defines the wrist camera frame $\{ wcf \}$ prior to teleoperation and keeps it fixed. The wrist camera frame definition procedure is as follows. For each axis of the wrist camera frame $\{ wcf \}$, the system asks the operator through a GUI to move their right wrist in a direction such that if the operator moves their wrist in the same direction during teleoperation, the end effector will move along the corresponding axis of the wrist camera frame $\{ wcf \}$. As such, the frame with which the desired end effector pose is represented wrt is aligned with the operator egocentric frame, eliminating the viewpoint mismatch issue.

More concretely, $\{ c \}$ represents the operator space camera frame. 
During the wrist camera frame definition procedure, the hand pose estimate predicts the positions of the operator right wrist with respect to (wrt) the camera frame $\{ c \}$. 
Our system thus obtains three sets of wrist positions, one for each axis of the wrist camera frame $\{ wcf \}$.
For each set of positions, we perform RANSAC line fitting to each set and then normalize the resulting lines to unit length. 
Let the normalized line corresponding to the $x$-axis of the wrist camera frame $\{ wcf \}$ be $n_x$. We similarly estimate $n_y, n_z$. 
We then project the matrix $\left[n_x, n_y, n_z\right] \in R^{3 \times 3}$ to the closet orthogonal matrix $R \in SO(3)$. 
We also compute the point $p$ closest to the three fitted lines $n_x, n_y, n_z$ in a least square sense. Let $\{ cf \}$ denotes the Cartesian operator frame. The procedure above produces $p$ and $R$, which together form the pose $^{c}T^{cf} = (p, R) \in SE(3)$ of the Cartesian operator frame $\{ cf \}$ wrt the camera frame $\{ c \}$.

\mypara{Obtaining motion correspondence between operator and robot.}
Given the pose of the Cartesian operator frame $^{c}T^{cf}$, we now describe how to compute the desired pose of the end effector given the current pose of the right hand wrist during teleoperation. Let $\{ wr \}$ denotes the frame attached to the operator right hand wrist. Let $\{ w \}, \{ e \}$ denote the robot world and end effector frames respectively. Given $ \prescript{c}{}{T}^{wr} $, we need to compute $ \prescript{w}{}{T}_{desired}^e $. We compute the desired position and orientation separately. Given the current position of the right hand wrist $ \prescript{c}{}{p}^{wr} $, we first express it wrt the Cartesian operator frame: $ \prescript{cf}{}{p}^{wr} = ^{cf}T^{c} \cdot \prescript{c}{}{p}^{wr}
$.

We then treat a scaled version of $\prescript{cf}{}{p}^{wr}$ to be the desired end effector position wrt the wrist camera frame $ \{ wcf \}$: $\prescript{wcf}{}{p}^{e} = \alpha \cdot \prescript{cf}{}{p}^{wr}$.

Since our system defines the frame $ \{ wcf \}$ prior to teleoperation and keeps the pose of $ \{ wcf \}$ wrt to the world frame $\{ w \}$ fixed during teleoperation, we can find the desired end effector position with:
\begin{align*}
    \prescript{w}{}{p}_{desired}^{e} = \prescript{w}{}{T}^{wcf} \cdot \prescript{wcf}{}{p}^{e}
\end{align*}
$\alpha$ is a scalar to control the scaling between human and robot motion. If $\alpha = 1$, the end effector moves the same distance as the operator wrist. The appropriate value of $\alpha$ depends on the task. For task like peg in hole, we might want to use a smaller $\alpha$ to allow for more precise robot control. 

Similarly, given the orientation $\prescript{c}{}{R}^{wr}$ of the operator right hand wrist frame $\{ wr \}$ wrt operator space camera frame $\{ c \}$, we can represent it wrt the Cartesian operator frame $\{ cf \}$ and then use the wrist camera frame $\{ wcf \}$ as a "bridge" to find the desired orientation of the end effector wrt the robot space world frame $\prescript{w}{}{R}_{desired}^{e}$. We elaborate on the rotational transformation procedure in the supplementary.

\mypara{Oblique operator frame.} To construct the Cartesian operator frame $\{ cf \}$, for each axis of the wrist camera frame $\{ wcf \}$, our system asks the operator through a GUI to move their right wrist in a direction such that if the operator moves their wrist in the same direction during teleoperation, the end effector will move along the same axis of the wrist camera frame $\{ wcf \}$. However, such promise by the system is often not possible because the $3$ best fit lines to the operator wrist positions are usually not orthogonal. This necessitates the projection step to $SO3$ as discussed above. However, projecting to $SO3$ the matrix whose columns are the three best-fit lines implies that as the operator moves their wrist in the same direction as one of the best fit lines, the robot end effector might not move along the corresponding axes of the wrist camera frame $\{ wcf \}$. In other words, if the operator repeats the same motion during the operator frame definition procedure and during teleoperation, the robot end effector might move in different directions. We found such non-repeatability of control commands unintuitive and therefore introduce the use of non-Cartesian oblique coordinate frame to ensure motion repeatability. 

\begin{figure}
\vspace{-2em}
\minipage{0.30\textwidth}
  \includegraphics[width=\linewidth]{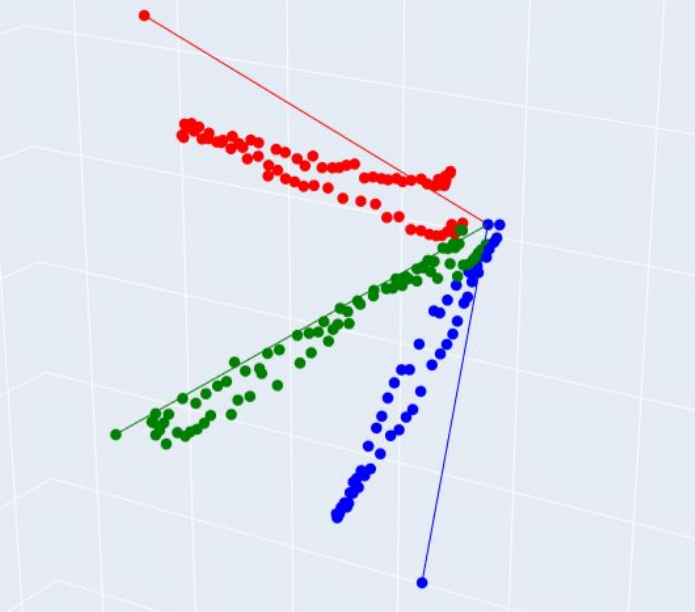}
\endminipage\hfill
\minipage{0.30\textwidth}
  \includegraphics[width=\linewidth]{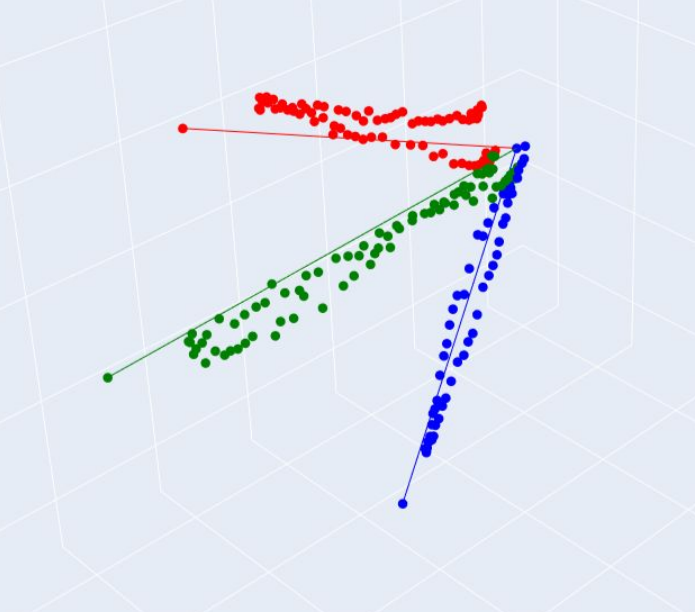}
\endminipage\hfill
\minipage{0.29\textwidth}
  \includegraphics[width=\linewidth]{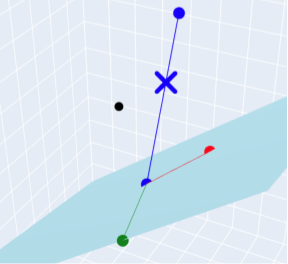}
\endminipage
  \caption{{\bf Left:} Cartesian operator frame. The z-axis of the Cartesian operator frame (blue line) is not parallel with the best fit lines of the blue points. {\bf Middle:} Oblique operator frame. {\bf Right:} Computing measure number in oblique frame: the black point is projected onto the z-axis, indicated by blue cross, along the direction parallel to the hyperplane defined by the x and y axes. }\label{fig:oblique_frame_illustration}
\vspace{-4mm}
\end{figure}

An oblique frame is a frame whose axes are not orthogonal~\cite{ObliqueCoord}. The difference between Cartesian and oblique coordinate frames are further illustrated in \autoref{fig:oblique_frame_illustration}. Given an arbitrary point in space, to find the measure number of a point wrt to an axis of an oblique coordinate frame, we project the point onto the axis along the direction parallel to the hyper-plane defined by the remaining axes. 

Let $\{ of \}$ denotes the oblique operator frame. We next explain how to obtain the oblique frame $\{ of \}$ from the cartesian frame $\{ cf \}$. We express the axes of the oblique coordinate frame as vector in the Cartesian operator frame. To find the x-axis of the oblique coordinate frame, we find the vector passing through the origin of the Cartesian operator frame and is parallel to the best fit lines of the 3d wrist positions obtained when defining the x-axis of the Cartesian operator frame. Denote the x-axis of the oblique frame in the Cartesian frame by $ \prescript{cf}{}{of}_x \in R^3$. Similarly, we can obtain the y-axis and z-axis of the oblique frame $\prescript{cf}{}{of}_y, \prescript{cf}{}{of}_z \in R^3$. 
The normal to the hyperplane defined by the x and y axes of the oblique frame is $ \prescript{cf}{}{n}_{xy} = \prescript{cf}{}{of}_x \times \prescript{cf}{}{of}_y$. 
Similarly, the normal the hyperplane defined by the remaining pairs of axes are $\prescript{cf}{}{n}_{xz}$ and $\prescript{cf}{}{n}_{yz}$.
Having defined these quantities, the computation to transform the 3d position $\prescript{c}{}{p}_x^{wr}$ of operator wrist to desired robot end effector position $\prescript{w}{}{p}_{desired}^{e}$ is:
\begin{align*}
\centering
    1. \prescript{cf}{}{p}^{wr} & = \prescript{cf}{}{T}^c . \prescript{c}{}{p}^{wr} \quad & 2. \prescript{of}{}{p}_x^{wr} & = \left< \prescript{cf}{}{p}^{wr}, \prescript{cf}{}{n}_{yz} \right> \Big/ \left< \prescript{cf}{}{of}_x , \prescript{cf}{}{n}_{yz} \right> \\ 
    3. \prescript{of}{}{p}_y^{wr} & = \left< \prescript{cf}{}{p}^{wr}, \prescript{cf}{}{n}_{xz} \right>  \Big/ \left< \prescript{cf}{}{of}_y , \prescript{cf}{}{n}_{xz} \right> \quad & 4.  \prescript{of}{}{p}_z^{wr} & = \left< \prescript{cf}{}{p}^{wr}, \prescript{cf}{}{n}_{xy} \right>  \Big/ \left< \prescript{cf}{}{of}_z , \prescript{cf}{}{n}_{xy} \right> \\
    5. \prescript{wcf}{}{p}^{e} & = \alpha . \left[ \prescript{of}{}{p}_x^{wr}, \prescript{of}{}{p}_y^{wr}, \prescript{of}{}{p}_z^{wr} \right] \in R^{3 \times 1} \quad & 6. \prescript{w}{}{p}_{desired}^{e} & = \prescript{w}{}{T}^{ wcf } . \prescript{ wcf }{}{p}^{e}
\end{align*}
, where $\left < \cdot, \cdot \right>$ is the inner product operator. $\alpha$ is the translational scaling factor as described above.

\subsection{Dynamic Motion Scaling}

\label{sec:dynamic_motion_scaling}

The translational motion scaling factor $\alpha$ determines the distance the end effector moves in response to movement by the operator. A smaller $\alpha$ allows for more precise control of the end effector. For example, if $\alpha = 0.02$, then the end effector will only move $2mm$ if the operator right wrist moves by $10cm$. This allows the operator to precisely control the position of the end effector without having to precisely control their wrist movement. Picking good values for $\alpha$ is thus important to successfully performing the tasks. However, good values for $\alpha$ change across tasks and even within one trial of a task depending on task progression. For example, in the peg in hole task, when the robot has not grasped the peg, a high value of $\alpha$ allows the operator to quickly command the robot to a good pre-grasp pose. However, when the peg is grasped and near the hole, a small value of $\alpha$ is preferable. 

Thus, when our teleoperation system only controls one robot arm, we use the estimated left hand poses to dynamically adjust the value of $\alpha$ during task execution. More concretely, if the y coordinate of the left hand wrist wrt the camera frame is below a certain threshold, we set $\alpha$ to $0.02$. Otherwise, we set $\alpha$ to $0.3$. In addition, if the distance in image space between the left hand thump and pinky tips is below a threshold, we set $\alpha$ to $0$. If $\alpha = 0$, there is no robot motion. 
Whenever $\alpha$ changes, we reinitialize the origin of the operator frame to be the current right wrist 3d position wrt camera frame.
This allows the operator to reset their position when they are almost out of the field of view of the camera. 
Dynamic motion scaling is an extremely useful yet simple technique that the operator takes advantage of extensively for difficult tasks such as peg in hole.

\vspace{-1mm}
\section{Experiments}
\vspace{-1mm}
\label{sec:problem}
\subsection{Task descriptions}

\begin{wrapfigure}{R}{.4\textwidth}
    \vspace{-0.5em}
    \begin{minipage}{\linewidth}
    \centering\captionsetup[subfigure]{justification=centering}
    \includegraphics[width=0.75\linewidth]{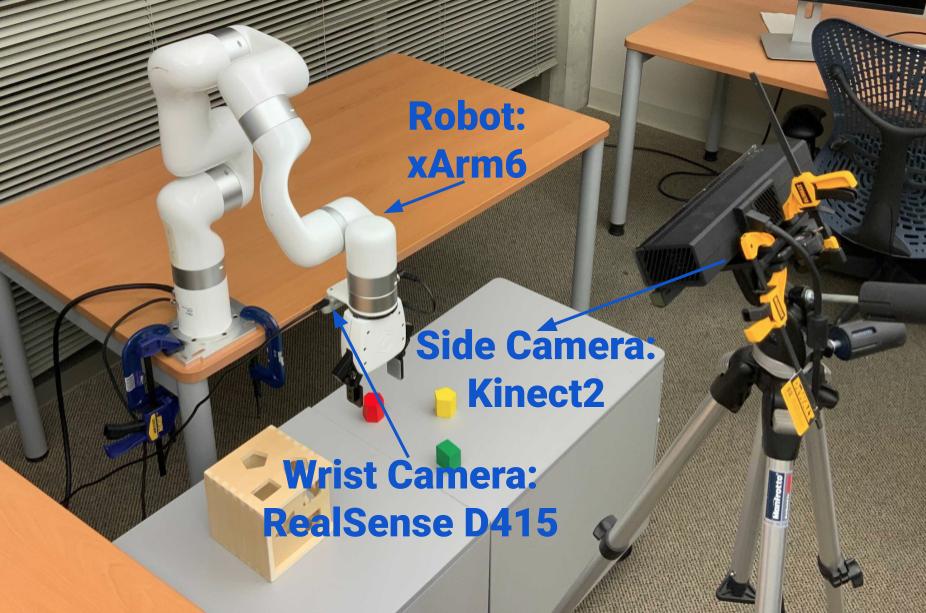}
    \subcaption{Hardware in robot space}
    \label{fig:hardware_robot_space}
    \includegraphics[width=0.75\linewidth]{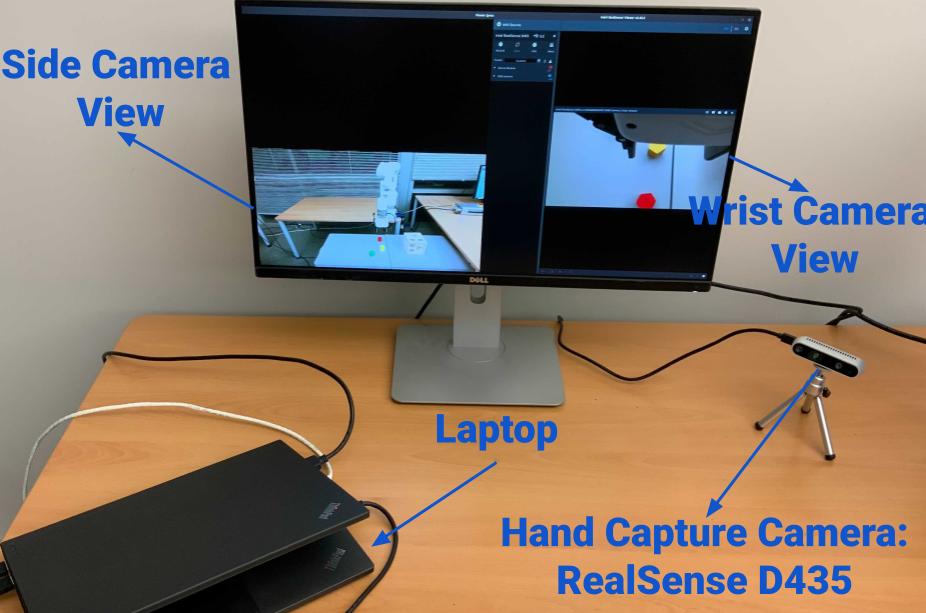}
    \subcaption{Hardware in operator space}
    \label{fig:hardware_operator_space}
    \vspace{-3em}
\end{minipage}
\end{wrapfigure}

The key question we investigate is whether given only a single RGB-D camera as the motion capture device, can a trained operator teleoperate the robot to perform complex manipulation tasks? To answer this question, we design a set of single arm manipulation tasks in real world or a dual arm coordination task in the simulator as shown in Figure~\ref{fig:teaser}. For the real robot experiments, we consider seven object manipulation tasks: pick and place, peg in hole, hammering, cutting fruit, folding cloth, cord untangling, and object storage in drawer. For the simulated experiment, we consider two-arm manipulation. We describe the tasks in more details below:

\mypara{Pick and Place} At the beginning of each episode, three typical objects from the YCB Dataset~\cite{calli2017yale} are placed in random positions on a table. The robot needs to grasp the object and move at least $0.6$ meters to reach the target position. The pick and place serves as the entry level manipulation task .

\mypara{Peg in Hole}
Three different types of pegs and holes are used in this task, ranked by difficulty from easy to hard: cylinder, pentagon, and hexagon. When the pegs are fully inserted in the holes, the clearance between them is no more than $3$ mm. We use this task to demonstrate the precision of our tele-operation system since a small position error will lead to a failure. 

\mypara{Hammering}
In this task, the robot first picks up a hammer with the gripper. The robot then needs to reach a small bench and hammer the wooden cylinder into the hole. This task evaluates whether the operator can grasp the hammer with a suitable pose and transmit force to the cylinder in the correct direction. Such tool manipulation tasks are of common interest in robotics community~\cite{billard2019trends, fang2020learning}. 

\mypara{Cutting Fruit}
Cutting fruit is another tool manipulation task. The robot uses a knife to slice a watermelon chunk into two pieces. Similar to hammering, cutting requires the robot to grasp the knife with the right pose and transmit force in the correct direction to the watermelon chunk.

\mypara{Folding Cloth}
Given a piece of cloth initially lying flat on a table, the robot folds the cloth twice along the diagonal. This task demonstrates the ability of our system to manipulate deformable object.

\mypara{Moving Large Container}
Two robot arms move a large container from one planar surface to another. Due to a lack of hardware, we demonstrate the successful completion of this task by our system in the simulator DRAKE~\cite{drake}. This task tests the coordination of two arms where each hand of the operator controls each robot arm. To control the second robot arm, we apply the same motion correspondence procedure described in \autoref{sec:method} to the left hand wrist.

\mypara{Cord Untangling}
One end of a cord is fixed. The remaining cord section is tangled to make a knot. The robot needs to untangle the cord by grasping and pulling the cord in the right direction.

\mypara{Object Storage in Drawer}
This task requires executing four primitive manipulation skills: pull the drawer to open, pick an object, place it into the drawer, and close the drawer. The task also requires manipulating articulated object, a problem which recently receives significant attention ~\cite{xiang2020sapien, urakami2019doorgym, katz2008manipulating}. 

For each tasks, the operator performs $3$ evaluation trials after a period of practice, except for \textit{Pick and Place} where we perform $5$ trials and \textit{Moving Large Container} where we only perform $1$ trial. 

We highlight that the \textit{Cord Untangling} and \textit{Object Storage in Drawer} tasks were not seen by the operator before evaluation. The first time the operator interacts with the objects present in these two tasks is during evaluation. Such unseen tasks allow us to test the adaptability of the system and the operator to new tasks and operating conditions. 

\subsection{Results of using RGB-D camera to capture hand poses}

  

\begin{wrapfigure}{R}{.35\textwidth}
    \vspace{-1.2em}
    \begin{minipage}{\linewidth}
    \centering\captionsetup[subfigure]{justification=centering}
    \includegraphics[width=0.9\linewidth]{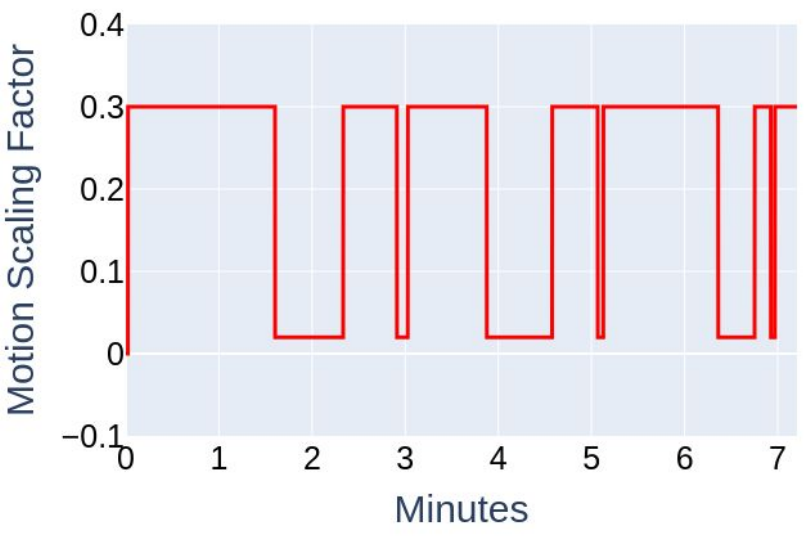}
    \subcaption{Peg in Hole}
    \label{fig:peg_in_hole}\par\vfill
    \includegraphics[width=0.9\linewidth]{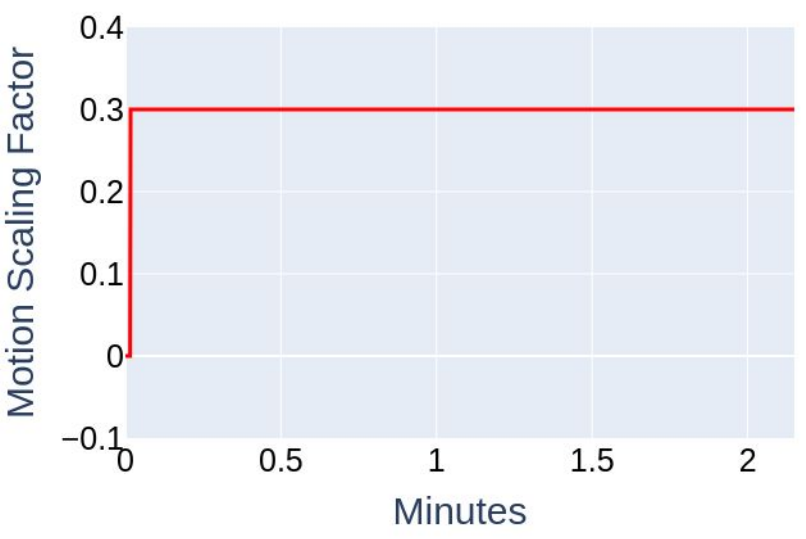}
    \subcaption{Pick and Place}
    \label{fig:pick_and_place}
\end{minipage}
\caption{In both figures, the x-axis is the duration of a trial and the y-axis is the dynamic translational motion scaling $\alpha$. {\bf Top:} For peg in hole, the operator switches back and forth between large and small $\alpha$ values depending on current progress {\bf Bottom:} For simple pick and place, the operator can successfully finish the task with a relatively high constant scaling factor.}\label{fig:alpha_fn_of_time} \vspace{-2em}
\end{wrapfigure}

For all the tasks, our teleoperation system successfully completes the evaluation trials. This fact is even more surprising for the \textit{Cord Untangling} and \textit{Object Storage in Drawer} tasks where the operator has not practiced with the objects before the evaluation trials. Such successes on unseen tasks demonstrate the generality of our teleoperation system and can potentially allow human operators to remotely control robot \textit{in the wild} in unseen conditions. The time to completion for each tasks are shown in Table \ref{table:mean_complete_time}. 

\begin{table}[t]
    \centering
    \centerline{\begin{tabular}{ |c||c|c|c|c|c|c|c| }
        \hline
         & Pick and & Peg in &  & Cutting  
            & Folding  & Object Storage  & Cord  \\
        Tasks & Place & Hole & Hammering & Fruit & Cloth & in Drawer & Untangling\\
        \hline
        Time & $8.9 \pm 1.3$ & $9.7 \pm 0.8$ & $3.8 \pm 0.8$ & $3.9 \pm 0.1$ 
            & $5.7 \pm 0.7$ & $7.0 \pm 1.2$ & $4.4 \pm 1.1$ \\
        \hline
    \end{tabular}}
    \vspace{1mm}
    \caption{Time and std to complete each task (in minutes). Our teleoperation system uses the end effector controller provided by the robot manufacturer. The controller has high latency and usually requires $2-3$ seconds to reach a desired end effector pose, leading to high task time-to-completion. Reducing the controller latency will significantly reduces the time-to-completion.}
    \label{table:mean_complete_time}
    \vspace{-6mm}
\end{table}


An interesting behavior we notice across tasks is failure recovery. For example in the first trial of the \textit{Cord Untangling} task, the end effector became stuck under a table in the robot space. Without manual reset of the robot, the operator was able to unstuck the end effector by moving under the table and eventually  complete the task. A common failure mode is closing the gripper without grasping the objects. This happens more frequently with small objects since neither the wrist camera or the side camera in the robot space provide distinctive visual cues to the operator whether the object is between the two grippers. However, this failure mode is easy to recover from because the operator can retry the grasp.

In addition to failure recovery, the operator also discovers interesting manipulation strategy such as dynamic manipulation. In the \textit{Cord Untangling} task when the end of the cord (with higher mass density) is stuck underneath itself, the operator quickly moves their wrist to induce a large position error, thus inducing a large acceleration of the robot, which unstucks the end of the cord due to inertia.

To ablate the benefit of the dynamic motion scaling scheme introduced in \autoref{sec:dynamic_motion_scaling}, the operator performs the peg in hole task while using a constant translational scaling factor $\alpha$. For a range of $\alpha$ values in $0.03, 0.3, 0.5, 1.0$, the operator was not able to successfully accomplish the task when given $15$ minutes to teleoperate the robot. For $\alpha=0.03$, the operator can not command the robot to reach the peg before going out of the field of view of the operator space camera. For higher values of $\alpha \in { 0.3, 0.5, 1.0 }$, the operator fails to control the robot precisely enough to insert the peg into the hole. The ability to dynamically modify the scaling factor $\alpha$ is most beneficial for precision manipulation tasks and less important for simple task, as illustrated in \autoref{fig:alpha_fn_of_time}.

\subsection{Results of using RGB camera to capture hand poses}

Using RGB camera instead of RGB-D camera is a promising extension to our method. The main challenge is to obtain accurate 3D positions of hand wrists without depth information. Monocular depth estimation is in general a very challenging problem. However, given that we only requires the 3D position of the right wrist, we were nevertheless able to obtain promising results. In addition, in our problem, the perception NN can have strong prior over the shape and size of the human hand, which improve the performance of 3D hand estimation over the general depth estimation problem. 

To estimate 3D hand pose from RGB, we crop the image around the hand region and predict the weak perspective transformation scale $s_h$ using the model from FrankMocap~\cite{rong2020frankmocap}. The weak perspective transformation approximates the true camera projection model. It assumes that for a small object, such as human hand, the depth value of any point on this object is the same. If the assumption holds, the pixel coordinates from weak projection well approximate the real projection model. We can thus approximate the depth simply by $\frac{Cf}{s_h}$, where $f$ is the focal length, $C$ is a scalar constant. 

Using only RGB to capture hand pose allows the operator to successfully complete $1$ out of $3$ trials of the pick and place task. In one other trial, the operator can grasp and move the three objects, but ends up knocking one of the objects over due to noisy depth estimate. In the remaining trial, monocular depth estimation fails when the right hand index and thump tips come close together to signal that the gripper should close. This violates the weak perspective assumption and leads to a non-trivial drop in the estimated depth. We had to stop the trial early. 


\vspace{-1mm}
\section{Conclusion}
\vspace{-1mm}
\label{sec:conclusion}
We propose a manipulation teleoperation system that uses a single RGB-D camera as the human motion capture device. The operator and our system can perform various complex tasks, even perform well on unseen ones. In the future, we are excited to expand the scopes of tasks that can be performed by our manipulation systems.





\bibliography{example}  

\end{document}